\documentclass[10pt,twocolumn,letterpaper]{article}

\usepackage{cvpr}
\usepackage{times}
\usepackage{epsfig}
\usepackage{graphicx}
\usepackage{amsmath}
\usepackage{amssymb}

\usepackage{relsize}
\usepackage{url}
\usepackage{tabularx, booktabs}
\usepackage{pifont}
\usepackage{enumitem}
\usepackage[table]{xcolor}
\definecolor{lightgray}{gray}{0.9}
\definecolor{lightblue}{rgb}{0.93,0.95,1.0}
\definecolor{darkgreen}{rgb}{0.0,0.6,0.0}
\definecolor{mypink1}{rgb}{0.858, 0.188, 0.478}

\newcommand{\T}{{\scriptscriptstyle \top}}
\newcommand{\loss}{\mathcal{L}}
\newcommand{\KL}{\mbox{KL}}

\newcommand{\minisection}[1]{\vspace{2mm}\noindent{\textbf{#1}.}}
\newcolumntype{L}[1]{>{\raggedright\let\newline\\\arraybackslash\hspace{0pt}}m{#1}}
\newcolumntype{C}[1]{>{\centering\let\newline\\\arraybackslash\hspace{0pt}}m{#1}}
\newcolumntype{R}[1]{>{\raggedleft\let\newline\\\arraybackslash\hspace{0pt}}m{#1}}

\newcommand{\secref}[1]{Sec.~\ref{#1}}
\newcommand{\siou}{Soft-IoU}
\newcommand{\diou}{Deep-IoU}
\newcommand{\base}{Base detector}
\newcommand{\scontext}{EM-Merger}
\newcommand{\dataset}{SKU-110K}
\newcommand{\cmark}{\textcolor{darkgreen}{\ding{51}}}
\newcommand{\xmark}{\textcolor{red}{\ding{55}}}

\def\beq{\begin{equation}}
\def\eeq{\end{equation}}

\def\beqary{\begin{eqnarray}}
\def\eeqary{\end{eqnarray}}

\def\beqarz{\begin{eqnarray*}}
\def\eeqarz{\end{eqnarray*}}

\usepackage[breaklinks=true,bookmarks=false]{hyperref}

\cvprfinalcopy 

\def\cvprPaperID{5953} 

\begin{document}
\title{Precise Detection in Densely Packed Scenes}
\author{
Eran Goldman$^{1,3\star}$ \,\,
Roei Herzig$^{2\star}$ \,\,
Aviv Eisenschtat$^{3\star}$ \,\,
Oria Ratzon$^3$ \,\,
Itsik Levi$^3$ \,\,\\
Jacob Goldberger$^1$ \,\,
Tal Hassner$^{4\dagger}$ \vspace{3pt}\\
$^1$Bar-Ilan University, $^2$Tel Aviv University, $^3$Trax Retail, $^4$The Open University of Israel}

\maketitle
\vspace*{-10mm}
\renewcommand*{\thefootnote}{$\star$}
\setcounter{footnote}{1}
\footnotetext{Equal Contribution.}
\renewcommand*{\thefootnote}{$\dagger$}
\setcounter{footnote}{1}
\footnotetext{Work done while at the University of Southern California.}
\renewcommand*{\thefootnote}{\arabic{footnote}}
\setcounter{footnote}{0}

\begin{abstract}
Man-made scenes can be densely packed, containing numerous objects, often identical, positioned in close proximity. We show that precise object detection in such scenes remains a challenging frontier even for state-of-the-art object detectors. We propose a novel, deep-learning based method for precise object detection, designed for such challenging settings. Our contributions include: (1) A layer for estimating the Jaccard index as a detection quality score; (2) a novel EM merging unit, which uses our quality scores to resolve detection overlap ambiguities; finally, (3) an extensive, annotated data set, \dataset, representing packed retail environments, released for training and testing under such extreme settings. Detection tests on \dataset{} and counting tests on the CARPK and PUCPR+ show our method to outperform existing state-of-the-art with substantial margins. The code and data will be made available on \url{www.github.com/eg4000/SKU110K_CVPR19}.
\end{abstract}


\begin{figure}[t!]
\centering
    \includegraphics[clip, trim=0mm 0mm 0mm 0mm,width=0.92\linewidth]{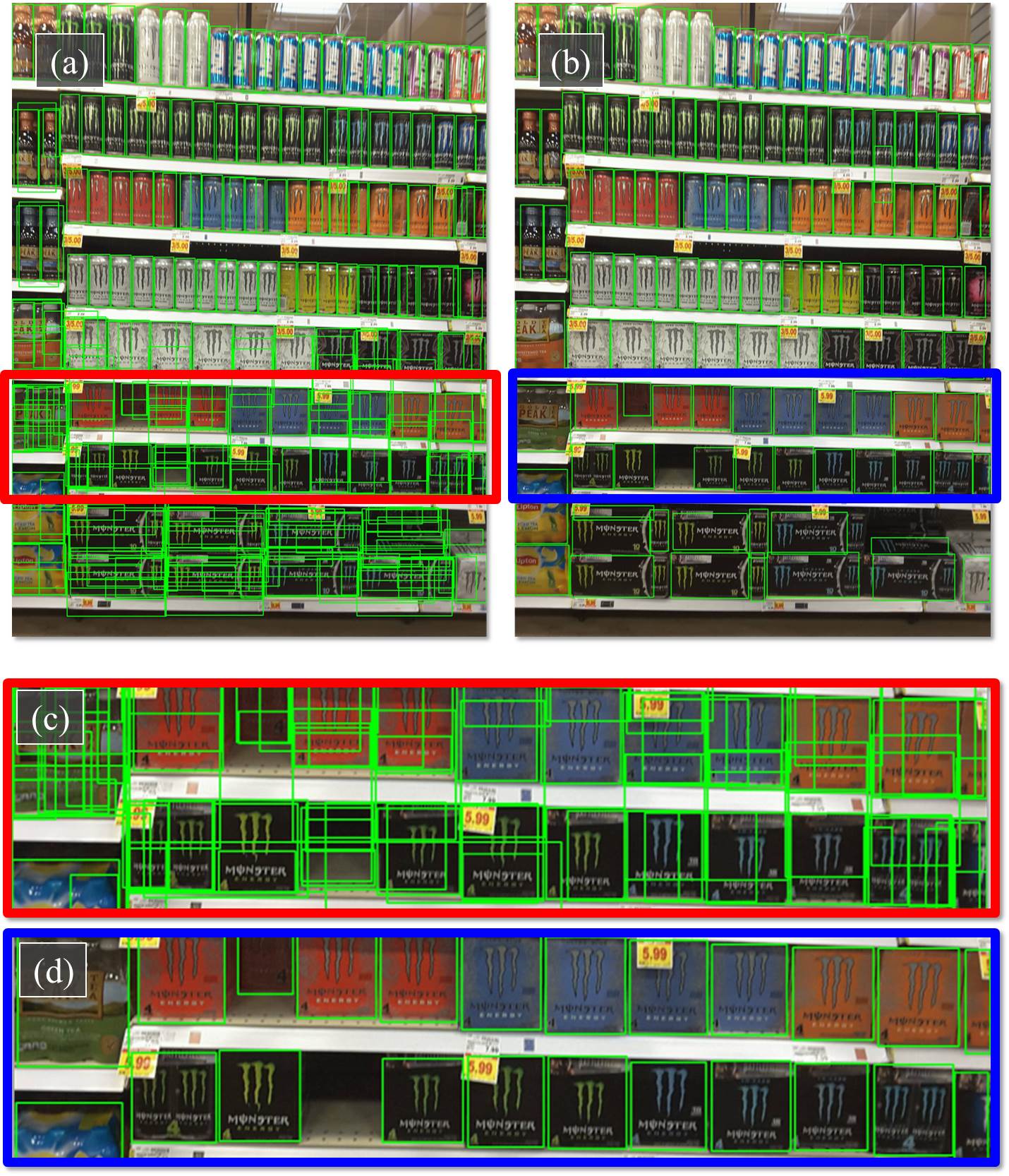}
\caption{{\bf Detection in packed domains.} A typical image in our \dataset{}, showing densely packed objects. (Top) (a) Detection results for the state-of-the-art RetinaNet~\cite{lin2018focal}, showing incorrect and overlapping detections, especially for the dark objects at the bottom which are harder to separate. (b) Our results showing far fewer misdetections and better fitting bounding boxes. (Bottom) Zoomed-in views for (c) RetinaNet~\cite{lin2018focal} and (d) our method. \vspace{-5mm}}
\label{fig:teaser}
\end{figure}

\section{Introduction}\label{sec:introduction}
Recent deep learning--based detectors can quickly and reliably detect objects in many real world scenes~\cite{rcnn,fast_rcnn,maskrcnn,lin2018focal,ssd,yolo,yolo9000,faster_rcnn}. Despite this remarkable progress, the common use case of detection in crowded images remains challenging even for leading object detectors.

We focus on detection in such {\em densely packed} scenes, where images contain many objects, often looking similar or even identical, positioned in close proximity. These scenes are typically man-made, with examples including retail shelf displays, traffic, and urban landscape images. Despite the abundance of such environments, they are underrepresented in existing object detection benchmarks. It is therefore unsurprising that state-of-the-art object detectors are challenged by such images.

\begin{table*}[t]
\centering
\normalsize
\resizebox{0.84\linewidth}{!}{
\rowcolors{1}{}{lightblue}
\begin{tabular}{lrrrrccc}
\toprule
{\bf Name} & {\bf \#Img.} & {\bf \#Obj./img.} & {\bf \hspace{8mm} \#Cls.} & {\bf \#Cls./img.} & {\bf \hspace{4mm} Dense.} & {\bf  \hspace{0mm}
 Idnt.} & {\bf \hspace{0mm} BB}\\ \hline
UCSD (2008)~\cite{chan2008privacy}    & 2000 & 24.9 & 1 & 1 & \cmark & \xmark & \xmark\\
PACAL VOC (2012)~\cite{pascalvoc}    & 22,531 & 2.71 & 20 & 2 & \xmark & \xmark & \cmark \\
ILSVRC Detection (2014)~\cite{imagenet_cvpr09}   & 516,840 & 1.12 & 200 & 2 & \xmark & \xmark  & \cmark\\
COCO (2015)~\cite{coco}   & 328,000 & 7.7 & 91 & 3.5 & \xmark & \xmark & \cmark \\
Penguins (2016)~\cite{arteta2016counting}  & 82,000 & 25 & 1 & 1 & \cmark & \xmark & \xmark\\
TRANCOS (2016)~\cite{onoro2016}    & 1,244 & 37.61 & 1 & 1 & \cmark & \cmark & \xmark \\
WIDER FACE (2016)~\cite{yang2016wider}    & 32,203 & 12 & 1 & 1 & \xmark & \xmark & \cmark \\
CityPersons (2017)~\cite{Shanshan2017CVPR}    & 5000 & 6 & 1 & 1 & \xmark & \xmark & \cmark \\
PUCPR+ (2017)~\cite{hsieh2017drone}    & 125 & 135 & 1 & 1 & \cmark & \cmark & \cmark \\
CARPK (2018)~\cite{hsieh2017drone}    & 1448 & 61 & 1 & 1 & \cmark & \cmark & \cmark \\
Open Images V4 (2018)~\cite{OpenImages}    & {\bf 1,910,098} & 8.4 & 600 & 2.3 & \xmark & \cmark & \cmark\\
{\bf Our \dataset{}} & {11,762} & {\bf 147.4} & {\bf 110,712} & {\bf 86} & \cmark & \cmark & \cmark  \\
\bottomrule
\end{tabular}
}
\caption{{\bf Key properties for related benchmarks.} {\bf \#Img.}: Number of images. {\bf \#Obj./img.}: Average items per image. {\bf \#Cls.}: Number of object classes (more implies a harder detection problem due to greater appearance variations).
{\bf \#Cls./img.}: Average classes per image. { \bf Dense}: Are objects typically densely packed together, raising potential overlapping detection problems? {\bf Idnt}: Do images contain multiple identical objects or hard to separate object sub-regions? {\bf BB}: Bounding box labels available for measuring detection accuracy?\vspace{-4mm}}
\label{tab:benchmarks}
\end{table*}

To understand what makes these detection tasks difficult, consider two identical objects placed in immediate proximity, as is often the case for items on store shelves (Fig.~\ref{fig:teaser}). The challenge is to determine where one object ends and the other begins; minimizing overlaps between their adjacent bounding boxes. In fact, as we show in Fig.~\ref{fig:teaser}(a,c), the state-of-the-art RetinaNet detector~\cite{lin2018focal}, often returns bounding boxes which partially overlap {\em multiple objects} or detections of adjacent object regions as separate objects.

We describe a method designed to accurately detect objects, even in such densely packed scenes (Fig.~\ref{fig:teaser}(b,d)). Our method includes several innovations. We propose learning the Jaccard index with a {\em soft Intersection over Union (\siou{})} network layer. This measure provides valuable information on the quality of detection boxes. We explain how detections can be represented as a {\em Mixture of Gaussians} (MoG), reflecting their locations and their \siou{} scores. An Expectation-Maximization (EM) based method is then used to cluster these Gaussians into groups, resolving detection overlap conflicts.

To summarize, our novel contributions are as follows:
\begin{itemize}
\item {\bf \siou{} layer}, added to an object detector to estimate the Jaccard index between the detected box and the (unknown) ground truth box (\secref{sec:method:siou}).
\item {\bf \scontext{} unit}, which converts detections and \siou{} scores into a MoG, and resolves overlapping detections in packed scenes (\secref{sec:scontext}).
\item {\bf A new data set and benchmark}, the store keeping unit, 110k categories (\dataset{}), for item detection in store shelf images from around the world (\secref{sec:db}). 
\end{itemize}
We test our detector on \dataset{}. Detection results show our method to outperform state-of-the-art detectors. We further test our method on the related but different task of {\em object counting}, on \dataset{} and the recent CARPK and PUCPR+ car counting benchmarks~\cite{hsieh2017drone}. Remarkably, although our method was not designed for counting, it offers a considerable improvement over state-of-the-art methods.


\section{Related work}\label{sec:related}
\minisection{Object detection} Work on this problem is extensive and we refer to a recent survey for a comprehensive overview~\cite{liu2018deep}. Briefly, early detectors employed sliding window--based approaches, applying classifiers to window contents at each spatial location~\cite{dalal2005histograms,felzenszwalb2010cascade,viola2004robust}. Later methods narrow this search space by determining {\em region proposals} before applying sophisticated classifiers~\cite{alexe2012measuring,carreira2012cpmc,rahtu2011learning,uijlings2013selective,zitnick2014edge}.

Deep learning--based methods now dominate detection results. To speed detection, proposal-based detectors such as R-CNN~\cite{rcnn} and Fast R-CNN~\cite{fast_rcnn} were developed, followed by Faster R-CNN~\cite{faster_rcnn} which introduced a {\em region proposal network} (RPN), then accelerated even more by R-FCN~\cite{Dai2016RFCNOD}. Mask-RCNN~\cite{maskrcnn} later added segmentation output and better detection pooling~\cite{faster_rcnn}. We build on these methods, claiming no advantage in standard object detection tasks. Unlike us, however, these {\em two-stage} methods were not designed for crowded scenes where small objects appear in dense formations. 

Recently, some offered {\em proposal-free} detectors, including YOLO~\cite{yolo}, SSD~\cite{ssd}, and YOLO9000~\cite{yolo9000}. To handle scale variance, feature pyramid network (FPN)~\cite{lin2017feature} added up-scaling layers. RetinaNet~\cite{lin2018focal} utilized the same FPN model, introducing a {\em Focal Loss} to dynamically weigh hard and easy samples for better handling of class imbalances that naturally occur in detection datasets. We extend this approach, introducing a new detection overlap measure, allowing for precise detection of tightly packed objects. 

These methods use hard-labeled log-likelihood detections to produce confidences for each candidate image region. We additionally predict a \siou{} confidence score which represents detection bounding box accuracy. 

\begin{figure*}[t!]
\centering
    \includegraphics[clip, trim=0mm 0mm 0mm 0mm,width=0.95\linewidth]{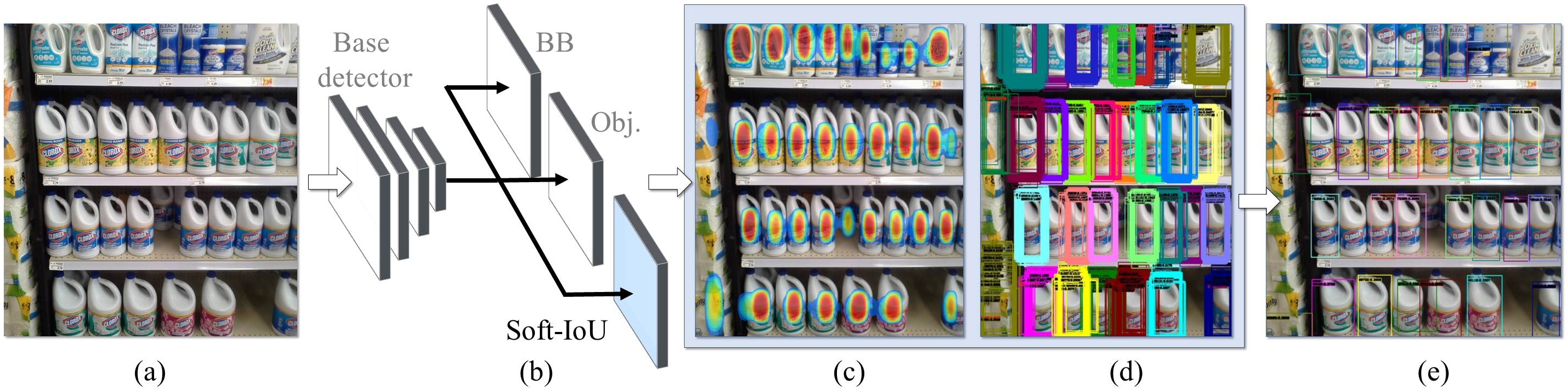}
\caption{{\bf System diagram.} (a) Input image. (b) A base network, with bounding box (BB) and objectness (Obj.) heads (Sec.~\ref{sec:method:model}), along with our novel \siou{} layer (Sec.~\ref{sec:method:siou}). (c) Our \scontext{} converts \siou{} to Gaussian heat-map representing (d) objects captured by multiple, overlapping bounding boxes. (e) It then analyzes these box clusters, producing a single detection per object~(e) (Sec.~\ref{sec:scontext}).\vspace{-3mm}}
\label{fig:system}
\end{figure*}

\minisection{Merging duplicate detections} Standard {\em non-maximum suppression} (NMS) remains a \textit{de-facto} object detection duplicate merging technique, from Viola \& Jones~\cite{viola2004robust} to recent deep detectors~\cite{lin2018focal, yolo9000,faster_rcnn}. NMS is a hand-crafted algorithm, applied at test time as post-processing, to greedily select high scoring detections and remove their overlapping, low confidence neighbors.

Existing NMS alternatives include mean-shift~\cite{dalal2005histograms,wojek2008sliding}, agglomerative~\cite{bourdev2010detecting}, and affinity propagation clustering \cite{mrowca2015spatial}, or heuristic variants~\cite{bodla2017soft,sermanet2013overfeat, iounet_eccv18}. GossipNet~\cite{hosang2017learning} proposed to perform duplicate-removal using a learnable layer in the detection network. Finally, others bin IoU values into five categories~\cite{tychsen2017improving}. We instead take a probabilistic interpretation of IoU prediction and a very different general approach.

Few of these methods showed improvement over simple, greedy NMS, with some also being computationally demanding~\cite{hosang2017learning}. In densely packed scenes, resolving detection ambiguities is exacerbated due to the many overlapping detections. We propose an unsupervised method, designed for clustering duplicate detection in cluttered regions.

\minisection{Crowded scene benchmarks} Many benchmarks were designed for testing object detection or counting methods and we survey a few in Table~\ref{tab:benchmarks}. Importantly, we are unaware of detection benchmarks intended for densely packed scenes, such as those of interest here. 

Popular object detection sets include ILSVRC~\cite{imagenet_cvpr09}, PASCAL VOC~\cite{pascalvoc} detection challenges, MS COCO~\cite{coco}, and the very recent Open Images v4~\cite{OpenImages}. None of these provides scenes with packed items. A number of recent benchmarks emphasize crowded scenes, but are designed for counting, rather than detection~\cite{arteta2016counting,chan2008privacy,onoro2016}.

As evident from Table~\ref{tab:benchmarks}, our new \dataset{} dataset, described in Sec.~\ref{sec:db}, provides {\em one to three orders of magnitude more items per image} than nearly all these benchmarks (the only exception is the PUCPR+~\cite{hsieh2017drone} which offers two orders of magnitude {\em fewer} images, and a single object class to our more than 110k classes). Most importantly, our enormous, per image, object numbers imply that all our images contain very crowded scenes, which raises the detection challenges described in Sec.~\ref{sec:introduction}. Moreover, identical or near identical items in \dataset{} are often positioned closely together, making detection overlaps a challenge. Finally, the large number of classes in \dataset{} implies appearance variations which add to the difficulty of this benchmark, even in challenges of object/non-object detection.

\section{Deep IoU detection network}\label{sec:method}
Our approach is illustrated in Fig.~\ref{fig:system}. We build on a standard detection network design, described in Sec.~\ref{sec:method:model}. We extend this design in two ways. First, we define a novel \siou{} layer which estimates the overlap between predicted bounding boxes and the (unknown) ground truth (Sec.~\ref{sec:method:siou}). These \siou{} scores are then processed by a proposed \scontext{} unit, described in Sec.~\ref{sec:scontext}, which resolves ambiguities between overlapping bounding boxes, returning a single detection per object.

\subsection{Base detection network}
\label{sec:method:model}
Our base detector is similar to existing methods~\cite{lin2017feature,lin2018focal,ssd,faster_rcnn}. We first detect objects by building a FPN network~\cite{lin2017feature} with three upscaling-layers, using ResNet-50~\cite{resnet1} as a backbone. The proposed model provides three fully-convolutional output heads for each RPN~\cite{faster_rcnn}: Two heads are standard and used also by previous work~\cite{lin2018focal,yolo9000} (our novel third head is described in Sec.~\ref{sec:method:siou}). 

The first is a {\em detection head} which produces a bounding box regression output for each object, represented as 4-tuples:~$(x,y,h,w)$ for the 2D coordinates of a bounding box center, height and width. The second, {\em classification head} provides an {\em objectness} score (confidence) label, $c\in[0,1]$ (assuming an object/no-object detection task with one object class). In practice, we filter detections for which $c\leq 0.1$, to avoid creating a bias towards noisy detections when training our \siou{} layer, described next.

\subsection{\siou{} layer}
\label{sec:method:siou}
In non-dense scenes, greedy NMS applied to objectness scores, $c$, can resolve overlapping detections. In dense images, however, multiple overlapping bounding boxes often reflect multiple, tightly packed objects, many of which receive high objectness scores. As we later show (Sec.~\ref{sec:exp:detection}), in such cases, NMS does not adequately discriminate between overlapping detections or suppress partial detections. 

\begin{figure}[t!]
  \begin{center}
  \centerline{\includegraphics[width=1\linewidth]{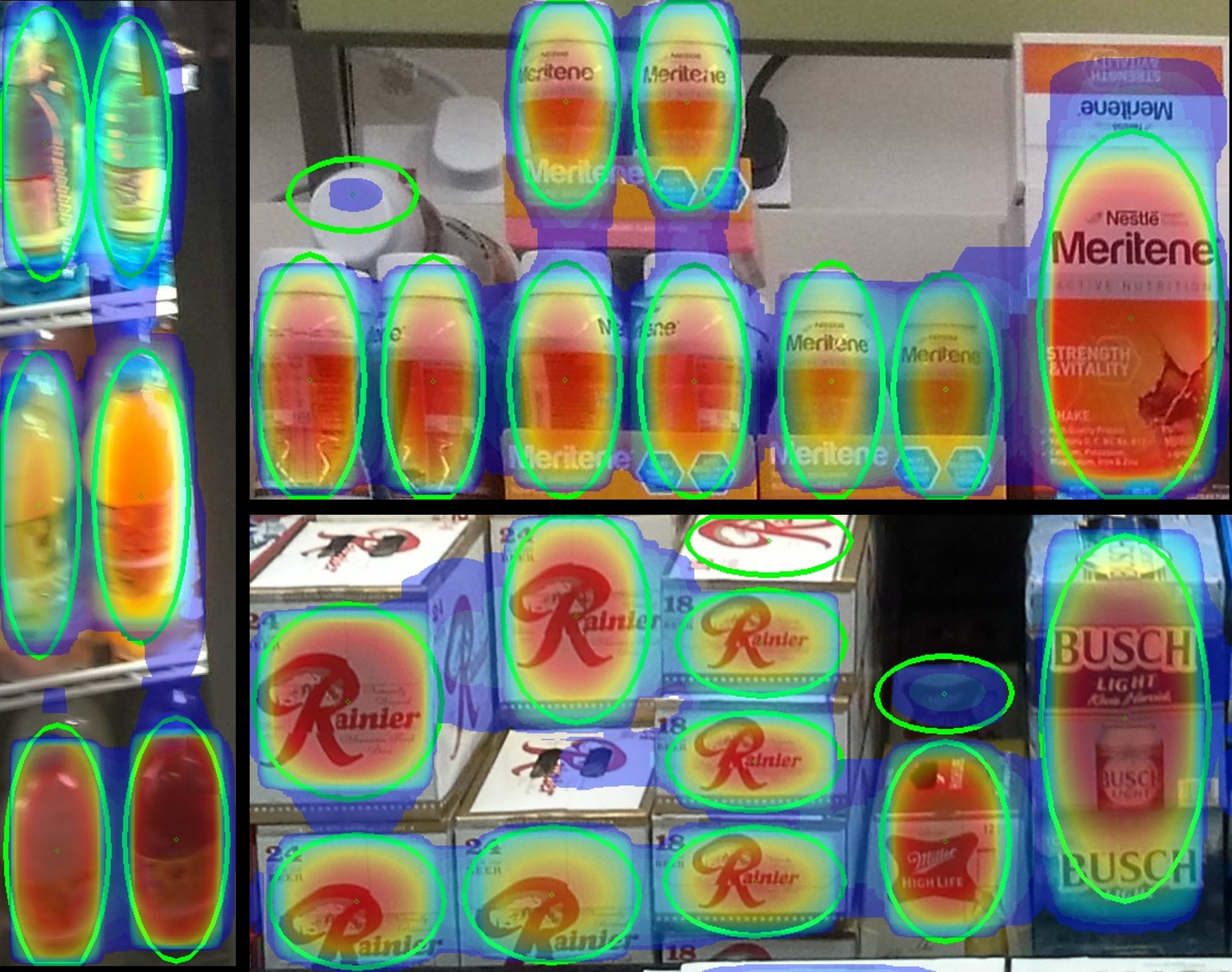}}
  \caption{{\bf Visualizing the output of the \scontext{} unit.} Raw detections on these images (not shown) contain many overlapping bounding boxes. Our approach of representing detections as a MoG (Eq.~\eqref{eq:aggAll}), visualized here as heat maps, provides clear signals for where items are located. The simplified MoG of Eq.~\eqref{eq:refinedMoG} is visualized as green ellipsoids. See Sec.~\ref{sec:scontext} for details.\vspace{-8mm}}
  \label{fig:GMM}
  \end{center}
\end{figure}

To handle these cluttered positive detections, we propose predicting an additional value for each bounding box: The IoU (\ie, Jaccard index) between a regressed detection box and the object location. This \siou{} score,~$c^{iou}\in[0,1]$, is estimated by a fully-convolutional layer which we add as a third head to the end of each RPN in the detector.

Given $N$ predicted detections, the IoU between a predicted bounding box $\mathbf{b}_{i}$, $i\in\{1..N\}$ and its ground truth bounding box, $\mathbf{\hat{b}}_{i}$, is defined as:
\begin{align}
IoU_i= \frac{Intersection(\mathbf{\hat{b}}_{i}, \mathbf{b}_{i})}{Union(\mathbf{\hat{b}}_{i}, \mathbf{b}_{i})}.\label{eq:IOU}
\end{align}
We chose $\mathbf{\hat{b}}_{i}$ to be the closest annotated box to $\mathbf{b}_{i}$ (in image coordinates). If the two do not overlap, then $IoU_{i}=0$. Both $Intersection(\cdot)$ and $Union(\cdot)$ count pixels.

We take a probabilistic interpretation of Eq.~\eqref{eq:IOU}, learning it with our \siou{} layer using a binary cross-entropy loss:
\begin{align}
\loss_{\text{sIoU}}&=& \label{eq:siou}\\
&-\frac{1}{n} \sum_{i=1}^n[IoU_{i}\log{(c_i^{iou})} + (1\!-\!IoU_{i})\log{(1\!-\!c_i^{iou})}],\nonumber
\end{align}
where $n$ is the number of samples in each batch.

The loss used to train each RPN in the detection network is therefore defined as:
\begin{equation}
\loss = \loss_{\text{Classification}} + \loss_{\text{Regression}} + \loss_{\text{sIoU}}.
\label{eq:total_loss}
\end{equation}
Here, $\loss_{\text{Classification}}$ and $\loss_{\text{Regression}}$ are the standard cross-entropy and euclidean losses, respectively~\cite{fast_rcnn,yolo,faster_rcnn}, and $\loss_{\text{sIoU}}$ is defined in Eq.~\eqref{eq:siou}. 

\minisection{Objectness vs. \siou{}} The objectness score used in previous methods predicts object/no-object labels whereas our \siou{} predicts the IoU of a detected bounding box and its ground truth. So, for instance, a bounding box which partially overlaps an object can still have a high objectness score, $c$, signifying high confidence that the object appears in the bounding box. For the same detection, we expect $c^{iou}$ to be low, due to the partial overlap. 

In fact, object/no-object classifiers are trained to be invariant to occlusions and translations. A good objectness classifier would therefore {\em be invariant} to the properties which our \siou{} layer is sensitive to. Objectness and \siou{} could thus be considered reflecting complementary properties of a detection bounding box.

\subsection{\scontext{} unit for inference}
\label{sec:scontext}
We now have $N$ predicted bounding box locations, each with its associated objectness, $c$, and \siou{}, $c^{iou}$, scores. Bounding boxes, especially in crowded scenes, often {\em clump} together in clusters, overlapping each other and their item locations. Our \scontext{} unit filters, merges, or splits these overlapping detection clusters, in order to resolve a single detection per object. We begin by formally defining these detection clusters.

\minisection{Detections as Gaussians} We consider the $N$ bounding boxes produced by the network as a set of 2D Gaussians:
\begin{equation}
\mathbf{F}=\{f_i\}_{i=1}^N=\{\mathcal{N}(\mathbf{p};{\mathbf{\mu}}_i, {\Sigma}_i)\}_{i=1}^N,
\end{equation}
with $\mathbf{p}\in\mathbb{R}^2$, a 2D image coordinate. The $i$-th detection is thus represented by a 2D mean, the central point of the box, $\mathbf{\mu}_i=(x_i,y_i)$, and a diagonal covariance, ${\Sigma}_i=[(h_i/4)^2,0;,0,(w_i/4)^2]$, reflecting the box size,~$(h_i,w_i)$.

We represent these Gaussians, jointly, as a single Mixture of Gaussians (MoG) density: 
\begin{equation}
 f(\mathbf{p}) = \sum_{i=1}^N \alpha_i f_i(\mathbf{p}),\label{eq:aggAll}
\end{equation}
where the mixture coefficients, $\alpha_i = \dfrac{c_i^{iou}}{\sum_{k=1}^N c_k^{iou}}$, reflecting our confidence that the bounding box overlaps with its ground truth, are normalized to create a MoG. 

Fig.~\ref{fig:GMM} visualizes the density of Eq.~\eqref{eq:aggAll} as heat-maps, translating detections into spatial region maps representing our per-pixel confidences of detection overlaps; each region weighted by the accumulated \siou{}.

\minisection{Selecting predictions: formal definition} We next resolve our $N$ Gaussians (detections) into precise, non-overlapping bounding box detections by using a MoG clustering method~\cite{bruneau2010parsimonious,goldberger2008simplifying,goldberger2005hierarchical,zhang2007simplifying}. 

We treat the problem of resolving the final detections as finding a set of $K<<N$ Gaussians,
\begin{equation}
\mathbf{G}=\{g_j\}_{j=1}^K = \{\mathcal{N}(\mathbf{p};{\mathbf{\mu'}}_j, {\Sigma'}_j)\}_{j=1}^K
\end{equation}
such that when aggregated, the selected Gaussians approximate the original MoG distribution $f$ of Eq.~\eqref{eq:aggAll}, formed by all $N$ detections. That is, if $g$ is defined by
\begin{equation}
 g(\mathbf{p}) = \sum_{j=1}^K \beta_j g_j(\mathbf{p}),\label{eq:refinedMoG}
\end{equation}
then we seek a mixture of $K$ Gaussians, $\mathbf{G}$, for which 
\begin{equation}
d(f,g) = \sum_{i=1}^N \alpha_i \min_{j=1}^K \KL ( f_i || g_j),
\label{dfg}
\end{equation}
is minimized, where $\KL$ is the KL-divergence~\cite{kullback1951information} used as a non-symmetric distance between two detection boxes.

\minisection{An EM-approach for selecting detections} We approximate a solution to minimization of Eq.~\eqref{dfg} using an EM-based method. The E-step assigns each box to the nearest box cluster, where box similarity is defined by a $\KL$ distance between the corresponding Gaussians. E-step assignments are defined as:
\begin{equation}
\label{eq:estep}
\pi(i) = \arg \min_{j=1}^K \KL ( f_i || g_j).
\end{equation}
The M-step then re-estimates the model parameters by:
\begin{align}
& \beta_j   =  \sum_{i\in\pi^{-1}(j)} \alpha_i  \nonumber\\ 
& \mu'_{j}  =   \frac{1}{\beta_j} \sum_{i\in\pi^{-1}(j)}\alpha_i\mu_{i} \label{eq:mstep}\\
& \Sigma'_{j} =  \frac{1}{\beta_j}
\sum_{i\in\pi^{-1}(j)}\alpha_i\left( \Sigma_{i}+
(\mu_{i}-\mu'_{j})(\mu_{i}-\mu'_{j})^{\T}\right). \nonumber
\end{align}
Note that these matrix computations are fast in 2D space. Moreover, all our Gaussians represent axis-aligned detection and so they all have diagonal covariances. In such cases, the KL distance between two Gaussians has a simpler form which is even more efficient to compute.

General EM theory guarantees that the iterative process described in Eq.~\eqref{eq:estep}--\eqref{eq:mstep}, is monotonically decreasing in the value of Eq.~\eqref{dfg} and converging to a local minimum~\cite{EM_paper}. We determine convergence when the value of Eq.~\eqref{dfg} is smaller than $\epsilon_{EM}= 1e-10$. We found this process to nearly always converge within ten iterations and so we set a maximum number of iterations at that number. 

EM parameters are often initialized using fast clustering to prevent convergence to poor local minima. We initialize it with an agglomerative, hierarchical clustering~\cite{rokach2005clustering}, where each detection initially represents a cluster of its own and clusters are successively merged until $K$ clusters remain.

We note in passing that there have been several recent attempts to develop deep clustering methods~\cite{Xie_2015,yang_2017}. Such methods are designed for clustering high-dimensional data, training autoencoders to map input data into a low-dimensional feature space where clustering is easier. We instead use EM, as these methods are not relevant in our settings, where the original data is two-dimensional. 

\minisection{Gaussians as detections} Once EM converged, the estimated Gaussians represent a set of $K$ detections. As an upper bound for the number of detections, we use $K=size(\mathbf{I}) / (\mu_w\mu_h)$, approximating the amount of non-overlapping, mean-sized boxes that fit into the image. As post-processing, we suppress less confident Gaussians which overlap other Gaussians by more than a predefined threshold. This step can be viewed as model selection and it determines the actual number of detected objects, $K' \leq K$.

To extract the final detections, for each of the $K'$ Gaussians, we consider the ellipse at two standard deviations around its center, visualized in Fig.~\ref{fig:GMM} in green. We then search the original set of $N$ detections (Sec.~\ref{sec:method:model}) for those whose center, $\mu=(x,y)$, falls inside this ellipse. A Gaussian is converted to a detection window by taking the median dimensions of the detections in this set. 

\section{The \dataset{} benchmark}\label{sec:db}
We assembled a new labeled data set and benchmark containing images of supermarket shelves. We focus on such retail environments for two main reasons. First, to maximize sales and store real-estate usage, shelves are regularly optimized to present many items in tightly packed, efficient arrangements~\cite{bell2017silent,nordfalt2014insights}. Our images therefore represent extreme examples of dense environments; precisely the type of scenes we are interested in. 

Second, retail items naturally fall into product, brand, and sub-brand object classes. Different brands and products are designed to appear differently. A typical store can sell hundreds of products, thereby presenting a detector with many inter-class appearance variations. Sub-brands, on the other hand, are often distinguishable only by fine-grained packaging differences. These subtle appearance variations increase the range of nuisances that detectors must face (\eg, spatial transformations, image quality, occlusion). 

As we show in Table~\ref{tab:benchmarks}, \dataset{} is very different from existing alternatives in the numbers and density of the objects appearing in each image, the variability of its item classes, and, of course, the nature of its scenes. Example images from \dataset{} are provided in Fig.~\ref{fig:teaser},~\ref{fig:system}, and~\ref{fig:qualitative}.

\minisection{Image collection} \dataset{} images were collected from thousands of supermarket stores around the world, including locations in the United States, Europe, and East Asia. Dozens of paid associates acquired our images, using their personal cellphone cameras. Images were originally taken at no less than five mega-pixel resolution but were then JPEG compressed at one megapixel. Otherwise, phone and camera models were not regulated or documented. Image quality and view settings were also unregulated and so our images represent different scales, viewing angles, lighting conditions, noise levels, and other sources of variability.

Bounding box annotations were provided by skilled annotators. We chose experienced annotators over unskilled, Mechanical Turkers, as we found the boxes obtained this way were more accurate and did not require voting schemes to verify correct annotations~\cite{coco,su2012crowdsourcing}. We did, however, visually inspect each image along with its detection labels, to filter obvious localization errors.

\minisection{Benchmark protocols} \dataset{} images were partitioned into train, test, and validate splits. Training consists of 70\% of the images ($8,233$ images) and their associated~$1,210,431$ bounding boxes; 5\% of the images ($588$), are used for validation (with their~$90,968$ bounding boxes). The rest, $2,941$ images ($432,312$ bounding boxes) were used for testing. Images were selected at random, ensuring that the same shelf display from the same shop does not appear in more than one of these subsets. 

\minisection{Evaluation} We adopt evaluation metrics similar to those used by COCO~\cite{coco}, reporting the average precision (AP) at IoU=.50:.05:.95 (their {\em primary challenge metric}), AP at IoU=.75, AP$^{.75}$ (their {\em strict metric}), and average recall (AR)$^{300}$ at IoU=.50:.05:.95 ($300$ is the maximal number of objects). We further report the value sampled from the precision-recall curve at recall $=0.5$ for IoU=0.75 (P$^{R=.5}$).

The many, densely packed items in our images are reminiscent of the settings in counting benchmarks~\cite{arteta2016counting,hsieh2017drone}. We capture both detection and counting accuracy, by borrowing the error measures used for those tasks: If $\{{K'}_i\}_{i=1}^n$ is the predicted numbers of objects in each test image, $i \in [1,n]$, and $\{t_i\}_{i=1}^n$are the per image ground truth numbers, then the mean absolute error (MAE) is ${\frac{1}{n} \sum_{i}^{n} |{K'}_i - t_{i}|}$ and the root mean squared error (RMSE) is ${\sqrt{\frac{1}{n} \sum_{i}^{n} ({K'}_i - t_{i})^{2}}}$.

\begin{table}[t!]
\centering
\resizebox{0.73\linewidth}{!}{
\begin{tabular}{lcc}
\toprule
Method & FPS & DPS \\ \hline
Faster-RCNN (2015) \cite{faster_rcnn} & 2.37 & 93\\
YOLO9000 (2017) \cite{yolo9000}  & 5 & 317\\ 
RetinaNet (2018) \cite{lin2018focal}  & 0.5 & 162 \\ \hline
\base{}  & 0.5 & 162 \\ 
+ \siou{}  & 0.5 & 162 \\ 
+ \scontext{} (on the CPU) & 0.23 &  73 \\ 
\bottomrule
\end{tabular}
}
\caption{{\bf Detection runtime comparison on~\dataset{}.}\vspace{-7mm}}\label{tab:runtime}
\end{table}

\section{Experiments}\label{sec:experiment}

\subsection{Run-time analysis}\label{sec:exp:runtime}
Table~\ref{tab:runtime} compares average frames per second (FPS) and detections per second (DPS) for baseline methods and variations of our approach. Runtimes were measured on the same machine using an Intel(R) Core(TM) i7-5930K CPU @3.50GHz GeForce and a GTX Titan X GPU. 

Our base detector is modeled after RetinaNet~\cite{lin2018focal} and so their runtimes are identical. Adding our \siou{} layer does not affect runtime. \scontext{} is slower despite the optimizations described in Sec.~\ref{sec:scontext}, mostly because of memory swapping between GPU and CPU/RAM. Our initial tests suggest that a GPU optimized version will be nearly as fast as the base detector.

\begin{table}[t!]
\centering
\resizebox{1.0\linewidth}{!}{
\begin{tabular}{l@{~~}c@{~}c@{~}c@{~}c@{~}c@{~}c}
\toprule
Method & AP & AP$^{.75}$ & AR$^{300}$  & P$^{R=.5}$& MAE  & RMSE\\ \hline
Monkey & .000 & 0 & .010 & 0 & N/A & N/A\\ \hline
Faster-RCNN~\cite{faster_rcnn} & .045 & .010 & .066 & 0 & 107.46 & 113.42\\ 
YOLO9000$^{opt}$~\cite{yolo9000}  & .094 & .073 &  .111 & 0 & 84.166 & 97.809\\ 
RetinaNet~\cite{lin2018focal} & .455 & .389 & .530 & .544 & 16.584 & 30.702\\ \hline 
Base \& NMS & .413 & .384 & .484 & .491 & 24.962 & 34.382\\ 
\siou{} \& NMS & .418 & .386 & .483 & .492 & 25.394 & 34.729\\ 
Base \& \scontext{}  & .482  & .540 & .553 & .802 & 23.978 & 283.971\\ \hline
Our full approach  & {\bf .492} & {\bf .556} &{\bf .554} & {\bf .834} & {\bf 14.522} & {\bf 23.992}\\
\bottomrule
\end{tabular}%
}
\caption{{\bf Detection on~\dataset{}.} Bold numbers are best results.
}\label{tab:skuresults}
\end{table}

\begin{figure}[t!]
\centering
\includegraphics[width=.97\linewidth]{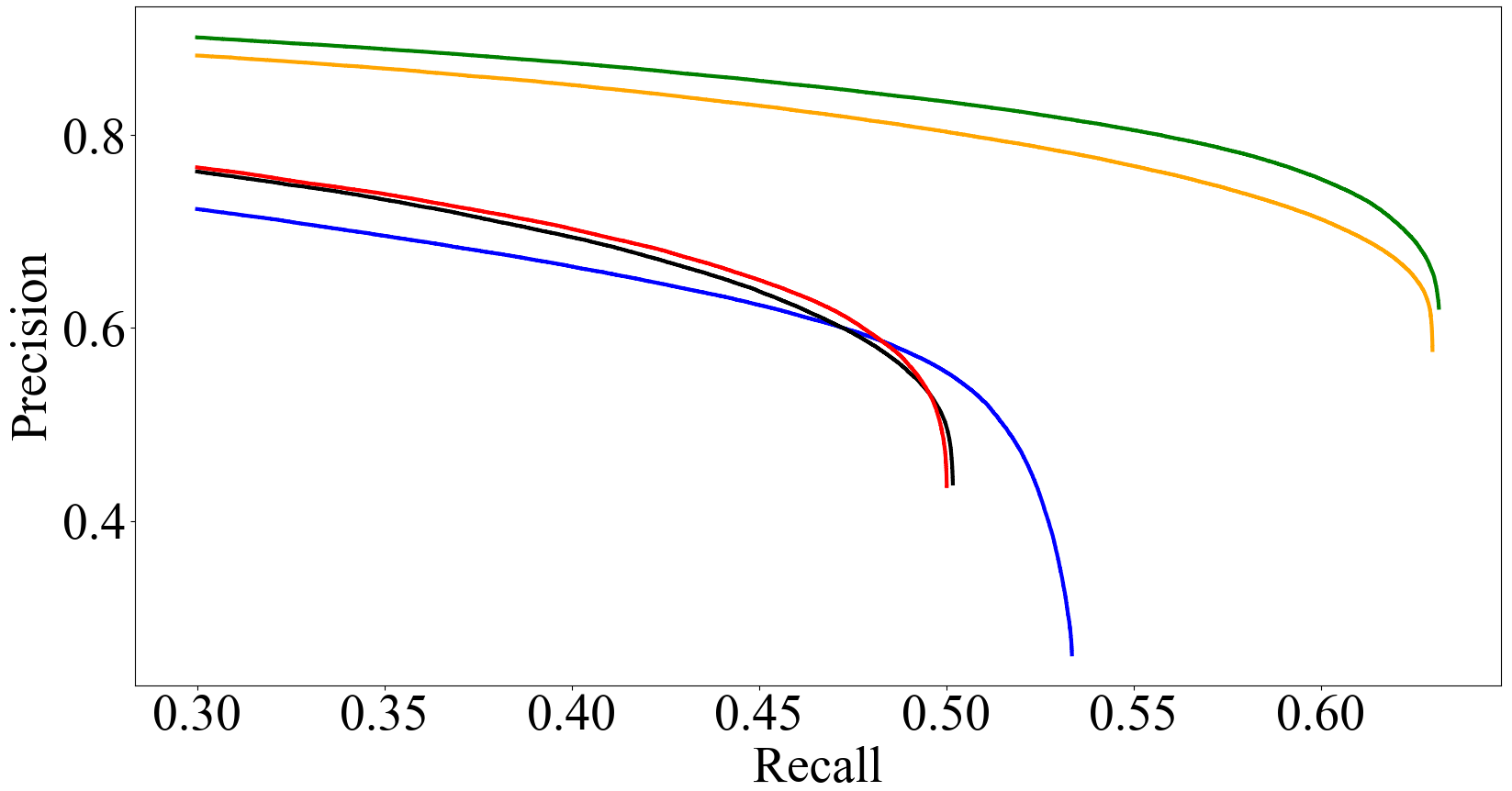}\\\hspace{5mm}
\includegraphics[width=.97\linewidth]{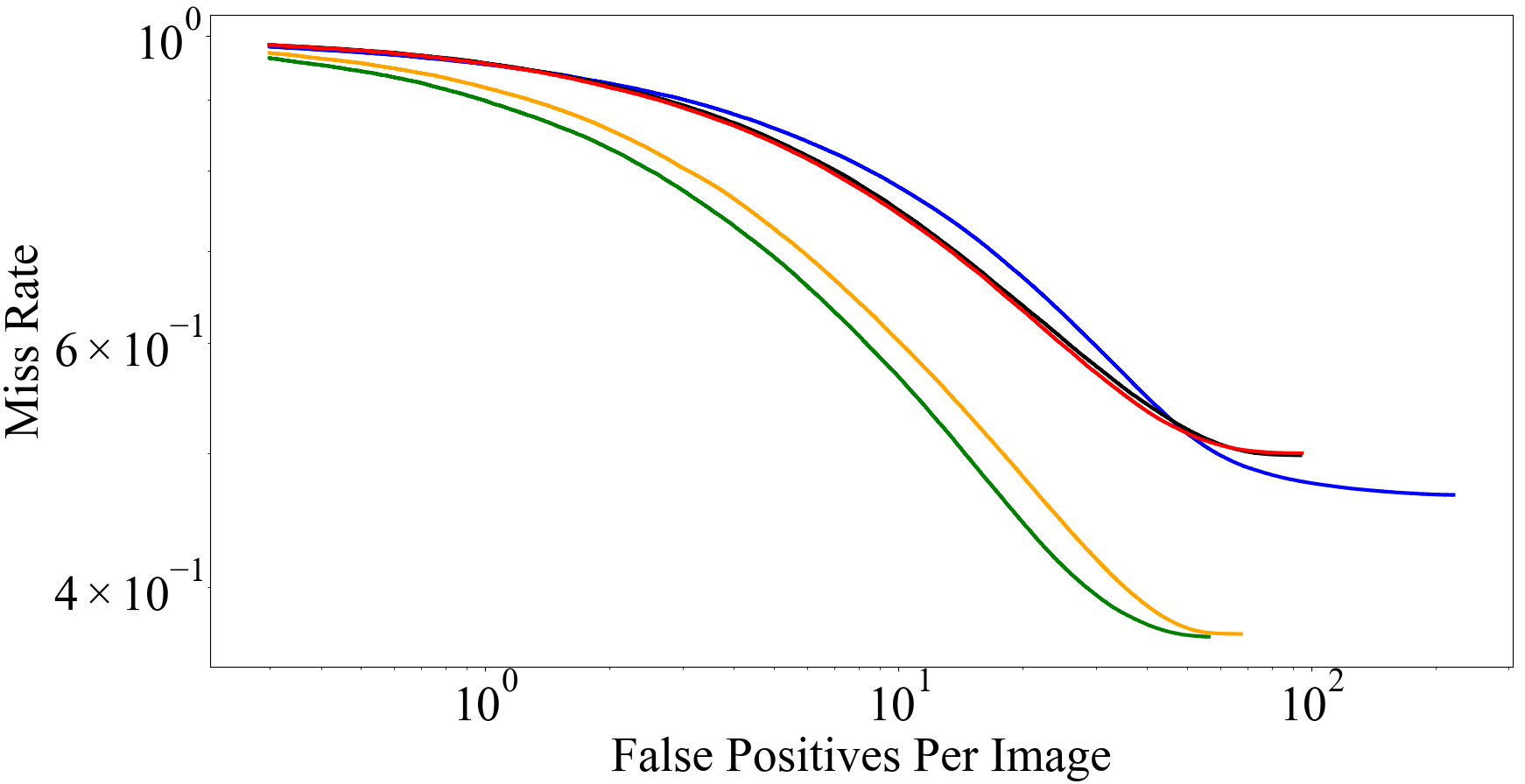}\\
\includegraphics[width=\linewidth]{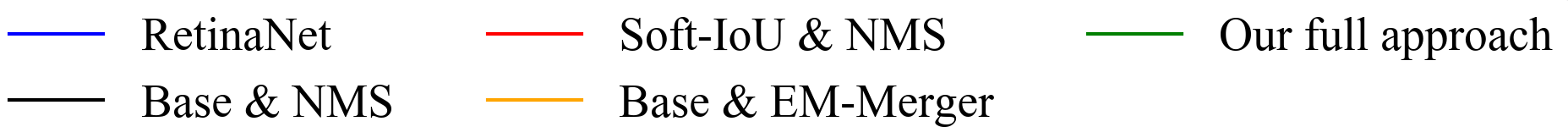}
\caption{{\bf Result Curves.} (a) PR Curves on~\dataset{} with IoU=0.75 ({\em higher} curve is better). (b) The log-log curve of miss rate vs False Positives Per Image~\cite{Shanshan2017CVPR} ({\em lower} curve is better).\vspace{-5mm}}
\label{fig:curves}
\end{figure}

\begin{figure*}[t!]
\centering
    \includegraphics[clip, trim=0mm 0mm 0mm 0mm,width=0.93\linewidth]{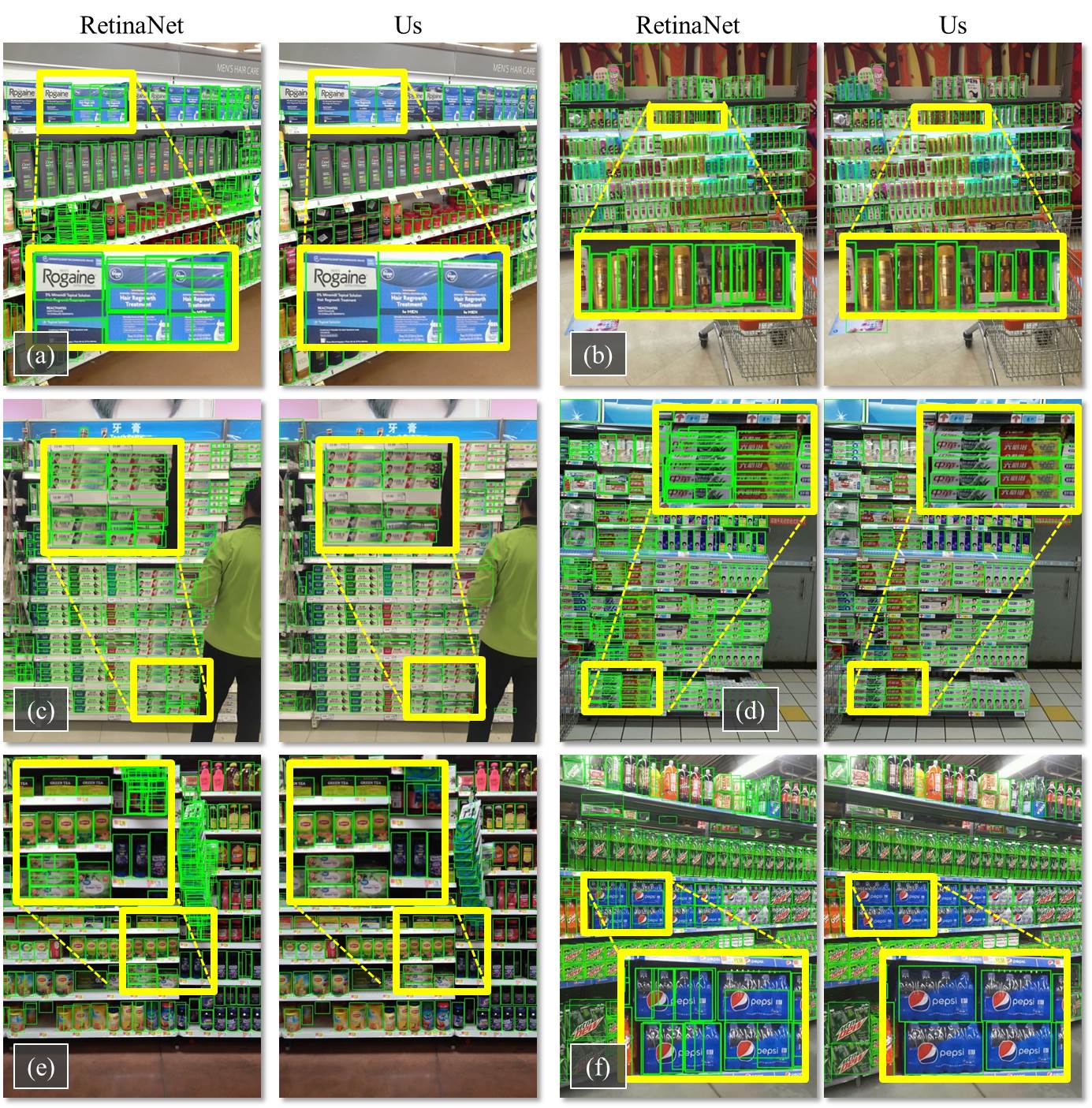}
\caption{{\bf Qualitative detection results on~\dataset{}.} Please see project web-page for more results and images in higher resolutions.\vspace{-3mm}}
\label{fig:qualitative}
\end{figure*}

\subsection{Experiments on the \dataset{} benchmark}
\label{sec:exp:detection}
\noindent{\bf Baseline methods.} We compare the detection accuracies of our proposed method and recent state-of-the-art on the \dataset{} benchmark. All methods, with the exception of the Monkey detector, were trained on the training set portion \dataset{}.

The following two baseline methods were tested using the original implementations released by their authors: {\bf RetinaNet}~\cite{lin2018focal} and {\bf Faster-RCNN}~\cite{faster_rcnn}. YOLO9000~\cite{yolo9000} is not suited for images with more than~50 objects. We offer results for {\bf YOLO9000$^{opt}$}, which is YOLO9000 with its loss function optimized and retrained to support detection of up to 300 boxes per image.

We also report the following ablation studies, detailing the contributions of individual components of our approach.
\begin{itemize}[nosep]
    \item {\bf Monkey:} Because of the tightly packed items in \dataset{} images, it is plausible that randomly {\em tossed} bounding boxes would correctly predict detections by chance. To test this naive approach, we assume we know the object number, $K'$, the mean and standard-deviation width, $\mu_w$, $\sigma_w$, and height, $\mu_h$, $\sigma_h$, for these boxes. Monkey samples 2D upper-left corners for the $K'$ bounding boxes from a uniform distribution and box heights and widths from Gaussian distributions $\mathcal{N}(h ;\mu_h,\sigma_h)$ and $\mathcal{N}(h;\mu_w,\sigma_w)$, respectively. 
    \item {\bf Base \& NMS}: Our basic detector of Sec.~\ref{sec:method:model} with standard NMS applied to objectness scores, $c$.
    \item {\bf \siou{} \& NMS}: Base detector with \siou{} (Sec.~\ref{sec:method:siou}). Standard NMS applied to \siou{} scores, $c^{iou}$, instead of objectness scores.
    \item {\bf Base \& \scontext{}}: Our basic detector, now using \scontext{} of Sec.~\ref{sec:scontext}, but applying it to original objectness scores, $c$.
    \item {\bf Our full approach}: Applying the \scontext{} unit to \diou{} scores, $c^{iou}$.
\end{itemize}

To test MAE and RMSE we report the number of detected objects, $K'$, and compare it with the true number of items per image. In {\em RetinaNet} the number of detections is extremely high so we first filter detections with low confidences. This confidence threshold was determined using cross-validation to optimize the results of this baseline.

\begin{table}[t!]
\centering
\resizebox{0.85\linewidth}{!}{
\begin{tabular}{lcc}
\toprule
Method & MAE  & RMSE \\ \hline
\multicolumn{3}{c}{{\bf Counting results on CARPK}} \\
Faster R-CNN (2015) \cite{faster_rcnn} & 24.32 & 37.62 \\ 
YOLO (2016) \cite{yolo} & 48.89 & 57.55 \\
One-Look Regression (2016) \cite{mundhenk2016large} & 59.46 & 66.84 \\ 
LPN Counting (2017) \cite{hsieh2017drone} & 23.80 & 36.79 \\ 
YOLO9000$^{opt}$ (2017) \cite{yolo9000} & 45.36 & 52.02 \\ 
RetinaNet (2018) \cite{lin2018focal} \cite{hsieh2017drone} & 16.62 & 22.30 \\ 
IEP Counting (2019) \cite{stahl2018divide} & 51.83 & - \\ \hline
Our full approach & {\bf 6.77} & {\bf 8.52} \\ \bottomrule

\multicolumn{3}{c}{{\bf Counting results on PUCPR+}} \\ 
Faster R-CNN (2015) \cite{faster_rcnn} & 39.88 & 47.67 \\ 
YOLO (2016) \cite{yolo} & 156.00 & 200.42 \\
One-Look Regression (2016) \cite{mundhenk2016large} & 21.88 & 36.73 \\ 
LPN Counting (2017) \cite{hsieh2017drone} & 22.76 & 34.46 \\ 
YOLO9000$^{opt}$ (2017) \cite{yolo9000} & 130.40 & 172.46 \\ 
RetinaNet (2018) \cite{lin2018focal} & 24.58 & 33.12 \\ 
IEP Counting (2019) \cite{stahl2018divide} & 15.17 & - \\ \hline
Our full approach & {\bf 7.16} & {\bf 12.00} \\ 
\bottomrule
\end{tabular}%
}
\caption{{\bf CARPK and PUCPR+ counting results}~\cite{hsieh2017drone}.\vspace{-9mm}}\label{tab:count}
\end{table}

\minisection{Detection results on \dataset{}} Quantitative detection results are provided in Table~\ref{tab:skuresults}, result curves are presented in Fig.~\ref{fig:curves}, and a selection of qualitative results, comparing our full approach with RetinaNet~\cite{lin2018focal}, the best performing baseline system, is offered in Fig.~\ref{fig:qualitative}.

Apparently, despite the packed nature of our scenes, randomly tossing detections fails completely, as evident by the near zero accuracy of Monkey. Both Faster-RCNN~\cite{faster_rcnn} and YOLO9000$^{opt}$~\cite{yolo9000} are clearly unsuited for detecting so many tightly packed objects. RetinaNet~\cite{lin2018focal}, performs much better, in fact outperforming our base network despite sharing a similar design (Sec.~\ref{sec:method:model}). This could be due to the better framework optimization of RetinaNet.

Our full system outperforms all its baselines with wide margins. Much of its advantage seems to come from our \scontext{} (Sec.~\ref{sec:scontext}). Comparing the accuracy of \scontext{} applied to either objectness scores or our \siou{} demonstrates the added information provided by \siou{}. This contribution is especially meaningful when examining the counting results, which show that \siou{} scores provide a much better means of filtering detection boxes than objectness scores.

It is further instructional to compare detection accuracy with counting accuracy. The counting accuracy gap between our method and the closest runner up, RetinaNet, is greater than the gap in detection accuracy (though both margins are wide). The drop in counting accuracy can at least partially be explained by their use of greedy NMS compared with our \scontext{}. In fact, Fig.~\ref{fig:qualitative} demonstrates the many overlapping and/or mis-localized detections produced by RetinaNet compared to the single detections per item predicted by our approach (see, in particular, Fig.~\ref{fig:qualitative}(a,e)).

Finally, we note that our best results remain far from perfect: The densely packed settings represented by \dataset{} images appear to be highly challenging, leaving room for further improvement.

\subsection{Experiments on CARPK and PUCPR+}\label{sec:exp:count}
We test our method on data from other benchmarks, to see if our approach generalizes well to other domains beyond store shelves and retail objects. To this end, we use the recent CARPK and PUCPR+~\cite{hsieh2017drone} benchmarks. Both data sets provide images of parking lots from high vantage points. We use their test protocols, comparing the number of detections per image to the ground truth numbers made available by these benchmarks. Accuracy is reported using MAE and RMSE, as in our \dataset~(Sec.~\ref{sec:db}).

\minisection{Counting results} We compare our method with results reported by others~\cite{hsieh2017drone, stahl2018divide}: {\bf Faster R-CNN}~\cite{faster_rcnn}, {\bf YOLO}~\cite{yolo}, and {\bf One-Look Regression}~\cite{mundhenk2016large}. Existing baselines also include two methods designed and tested for counting on these two benchmarks: {\bf LPN Counting}~\cite{hsieh2017drone} and {\bf IEP Counting}~\cite{stahl2018divide}. In addition, we trained and tested counting accuracy with {\bf YOLO9000$^{opt}$}~\cite{yolo9000} and {\bf RetinaNet}~\cite{lin2018focal}.

Table~\ref{tab:count} reports the MAE and RMSE for all tested methods. Despite not being designed for counting, our method is more accurate than recent methods designed for that task. A significant difference between these counting datasets and our \dataset{} is in the much closer proximity of the objects in our images. This issue has a significant impact on baseline detectors, as can be seen in Tables~\ref{tab:count} and~\ref{tab:skuresults}. Our model suffers a much lower degradation in performance due to better filtering of these overlaps.\footnote{See project web-page for qualitative results on these benchmarks.}

\section{Conclusions}\label{sec:conclusions}
The performance of modern object/no-object detectors on existing benchmarks is remarkable yet still limited. We focus on densely packed scenes typical of every-day retail environments and offer \dataset{}, a new benchmark of such retail shelf images, labeled with item detection boxes. Our tests on this benchmark show that such images challenge state-of-the-art detectors. 

To address these challenges, along with our benchmark, we offer two technical innovations designed to raise detection accuracy in such settings: The first is a \siou{} layer for estimating the overlap between predicted and (unknown) ground truth boxes. The second is an EM-based unit for resolving bounding box overlap ambiguities, even in tightly packed scenes where these overlaps are common. 

We test our approach on \dataset{} and two existing benchmarks for counting, and show it to surpass existing detection and counting methods. Still, even the best results on \dataset{} are far from saturated, suggesting that these densely packed scenes remain a challenging frontier for future work.

\section*{Acknowledgement}
This research was supported by \emph{Trax Image Recognition for Retail and Consumer Goods} \url{https://traxretail.com/}. We are thankful to Dr. Yair Adato and Dr. Ziv Mhabary for their essential support in this work.

{\small
\bibliographystyle{ieee}
\bibliography{trax}
}

\end{document}